\documentclass{article}
\usepackage{fancyhdr}
\usepackage{spconf,amsmath,graphicx}
\usepackage{caption}
\usepackage{subcaption}
\usepackage{multirow} 
\usepackage{xcolor}
\usepackage{svg}
\usepackage{mathrsfs}
\usepackage{dutchcal}
\usepackage{graphics}
\usepackage{comment}
\DeclareUnicodeCharacter{2212}{-}
\usepackage{hyphenat}

\usepackage{hyperref}

\title{SimHaze: game engine simulated data for real-world dehazing}

\makeatletter
\def\@name{\emph{Zhengyang Lou$^{\star, \dagger}$, Huan Xu$^{\star}$, Fangzhou Mu$^{\star}$,  Yanli Liu$^{\dagger}$}, \\ \emph{Xiaoyu Zhang$^{\dagger}$, Liang Shang$^{\star, \dagger}$, Jiang Li$^{\dagger}$, Bochen Guan$^{\dagger}$, Yin Li$^{\star}$, Yu Hen Hu$^{\star}$}}
\makeatother

\address{$^{\star}$ University of Wisconsin-Madison,
Madison, WI, 53706\\
$^{\dagger}$ OPPO US Research Center, Innopeak Technology Inc, Palo Alto, CA, 94303}

\fancypagestyle{FirstPage}{
\lfoot{\copyright 2023 IEEE. Personal use of this material is permitted. Permission from IEEE must be obtained for all other uses, in any current or future media, including reprinting/republishing this material for advertising or promotional purposes, creating new collective works, for resale or redistribution to servers or lists, or reuse of any copyrighted component of this work in other works.}
}

\begin{document}
%

\maketitle{}
\begin{abstract}
Deep models have demonstrated recent success in single-image dehazing. Most prior methods consider fully supervised training and learn from paired clean and hazy images, where a hazy image is synthesized based on a clean image and its estimated depth map. This paradigm, however, can produce low-quality hazy images due to inaccurate depth estimation, resulting in poor generalization of the trained models. In this paper, we explore an alternative approach for generating paired clean-hazy images by leveraging computer graphics. Using a modern game engine, our approach renders crisp clean images and their precise depth maps, based on which high-quality hazy images can be synthesized for training dehazing models. To this end, we present \textit{SimHaze}, a new synthetic haze dataset. More importantly, we show that training with \textit{SimHaze} alone allows the latest dehazing models to achieve significantly better performance in comparison to previous dehazing datasets. Our dataset and code will be made publicly available. 
\end{abstract}

\section{Introduction}
\label{sec:intro}
\thispagestyle{FirstPage}


Haze is a natural phenomenon due to the presence of dense airborne particles that causes unwanted distributed light scattering that typically happens in outdoor scenes. Dehazing is a necessary image enhancement procedure to mitigate image quality degradation due to the presence of haze. Single-image dehazing has been a long-standing challenge, with recent development focused on deep learning-based methods.

Many existing deep models follow a fully supervised approach, and learn from pairs of pixel-aligned clean and hazy images~\cite{song2022vision,guo2022dehamer,cai2016dehazenet,Ren-ECCV-2016,li2017all,10.1145/3474085.3475432,chen2018gated,qin2020ffa}. These paired images are difficult to collect at scale. Prior works hence resort to synthesizing hazy images using physics-based models, e.g., Atmospheric Scattering Model (\textit{ASM})~\cite{narasimhan2002vision}, where a hazy image is synthesized based on a clean image and its depth map, often predicted by monocular depth estimation~\cite{RESIDE,4KID}.

A key limitation of this paradigm is that the quality of synthesized hazy images heavily depends on the accuracy of depth maps provided by error-prone monocular depth estimation methods. This could lead to poor generalization of natural hazy scenes with a high level of ambiguity for monocular depth estimation. Indeed, a recent study~\cite{gui2021comprehensive} shows that supervised deep models fall short at outdoor scenes, where depth estimation is less accurate. 

An appealing solution is generating high-quality clean-hazy image pairs using computer graphics. A recent work (DLSU~\cite{DLSU}) considers rendering clean-hazy images using a game engine for training dehazing models. However, the authors conclude that training existing dehazing models using synthetic data produces unsatisfactory results, due to a major domain gap between the synthesized and the real images. They thus propose to design a special dehazing model.

In this paper, we present an orchestrated data generation pipeline to render synthetic images for training dehazing models. Contrary to DLSU~\cite{DLSU}, our key finding is that \textit{high photorealism} and \textit{sufficient randomization} of camera trajectories and haze parameters enable the training of existing dehazing models with superior performance. 

Specifically, we make use of the cutting-edge Unreal 4~\cite{unrealengine} game engine to generate high-quality images and precise depth maps. We then employ \textit{ASM} to synthesize hazy images at various lighting conditions, creating a comprehensive training dataset consistent with the commonly adopted RESIDE~\cite{RESIDE} benchmark. We conduct extensive experiments and demonstrate both qualitative and quantitative results. Our results suggest that dehazing algorithms trained on our dataset outperform those trained on existing hazy image datasets, when applied to natural hazy images.\medskip

\noindent \textbf{Contributions}. In summary, our contributions include: a) Designing and analyzing a workflow for simulating photo-realistic hazy images using a game engine. b) Demonstrating that existing deep models can achieve state-of-the-art results on real images with a surprisingly simple training procedure using our simulated data. c) Introducing \textit{SimHaze}, a large-scale synthetic dataset for training image dehazing models.

\begin{figure*}
\centerline{\includegraphics[width=500pt]{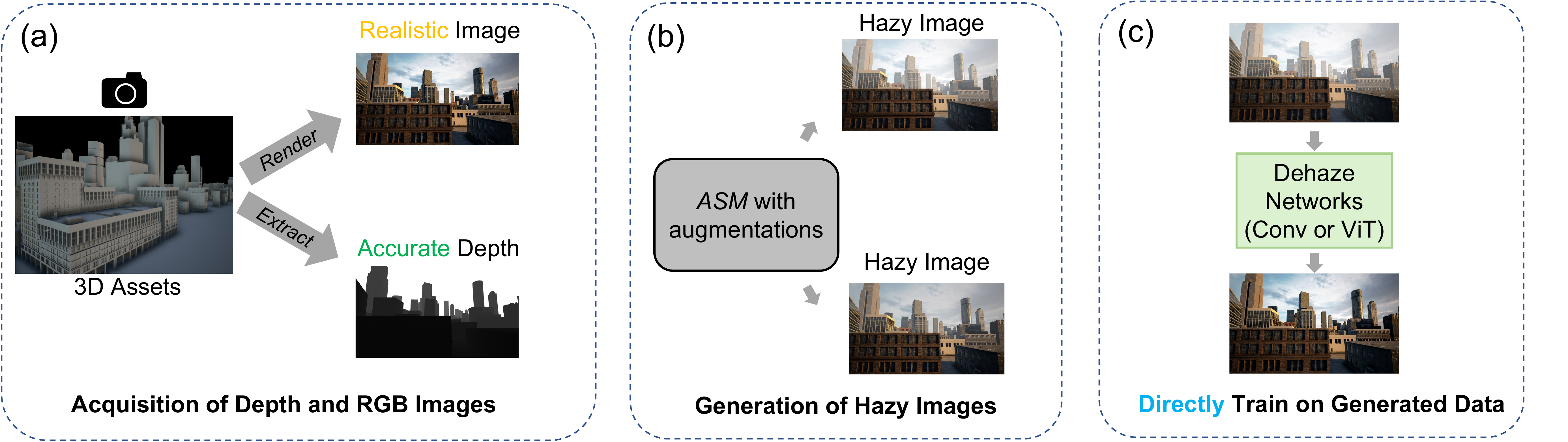}}
    \caption{Overview of our method. (a). We first render \textit{realistic} clean RGB images and \textit{accurate} depth maps from a modern game engine.  (b). We then synthesize training pairs with clean and hazy images based on a physical-based haze model, \textit{ASM} from RGB images, and depth maps from (a). (c). With high \textit{photorealism}, \textit{accurate depth} and \textit{sufficient randomization} of camera trajectories and haze parameters, dehazing models can be \textit{directly} trained on our dataset and produce competitive results.}
    \label{fig: pipeline}
\end{figure*}
\section{Background}

\subsection{Atmospheric Scattering Model}

Denote the observed intensity (hazy image) and the scene radiance (clean image) as $H(x,y)$ and $I(x,y)$, where $x$ and $y$ index pixel locations on a finite 2D grid defining the image plane. The relation between $H(x,y)$ and $I(x,y)$ can be described by the \textit{ASM}~\cite{narasimhan2002vision}:
\begin{equation}\label{ASM_equation}
     H(x,y) = t(x,y)I(x,y) + (1-t(x,y))A,
 \end{equation}
where $A$ is the global atmospheric light, $t(x,y)$ is the {\it transmission map}. $t(x,y)$ describes the percentage of light arriving at the camera without being scattered. $t(x,y)I(x,y)$ indicates how scene radiance decay over the haze. $(1-t(x,y))A$ comes from environment light being scattered into the location $(x,y)$, and is the main reason for the color distortion.

For a homogeneous medium, $t(x,y)$ follows:
\begin{equation}\label{trans}
     t(x,y) = e^{-\beta d(x,y)},
\end{equation}
where $\beta$ is the atmosphere scattering parameter and controls the density of the haze, and $d(x,y)$ is the depth map. Intuitively, the equation suggests that the scene radiance decays exponentially with the depth.


\subsection{Existing Dehazing Datasets}
Existing dehazing image datasets, such as RESIDE \cite{RESIDE} and 4KID \cite{4KID} are developed by first collecting a set of haze-free natural images $I(x,y)$. Then deep learning-based monocular depth estimation algorithms such as~\cite{liu2015learning} are applied to estimate the corresponding depth map $d(x,y)$. A set of corresponding hazy images $H(x,y)$ will be generated using Eq. (\ref{ASM_equation}) with randomly sampled values of the atmosphere scattering parameter $\beta$ and the global atmospheric light $A$. Multiple hazy images $H(x,y)$ corresponding to a given image $I(x,y)$ is generated and then used for training dehazing models. 


The above datasets estimate the depth map directly from a single image. DLSU \cite{DLSU}, on the other hand, uses rendered images from a 3D virtual environment as haze-free images. Then a domain adaptation step is applied. However, there is a major domain gap even after the adaptation, a special dehazing model is thus proposed that requires explicit estimation of the depth, transmission, and atmospheric maps. 
Unlike DLSU, \textit{SimHaze} does not require domain adaptation and can be used to train state-of-the-art dehazing models that do not explicitly estimate $A$ and $t$.  






\section{Method}\label{Method}

\subsection{Data Generation}

Our approach leverages the powerful Unreal Engine 4 \cite{unrealengine} to render realistic $I(x,y)$ and accurate depth map $d(x,y)$, thereby avoiding depth estimation errors. 
After obtaining clean images and depth maps, \textit{ASM} is used to synthesize hazy images. The synthetically rendered clean images and their corresponding hazy images are then used to train deep models for single-image haze removal. 
An overview is shown in Figure \ref{fig: pipeline}. In what follows we describe the details.\medskip 


\noindent \textbf{Acquisition of Depth and RGB Images}. Unreal exposes a highly flexible user interface for the acquisition of accurate depth information, which is the key to realistic haze synthesis. To automate the sampling of camera positions and orientations, we use Unreal's built-in Python Scripting API. With the camera poses sampled, depth and RGB images of the same scene can both be rendered by Unreal's built-in rendering system. This process is depicted by in Figure \ref{fig: pipeline}(a).\medskip

\noindent \textbf{Generation of Hazy Images}. After the clean RGB images and depth images are extracted, Eq.\ \ref{ASM_equation} is used to construct the hazy images. Following the outdoor training set of~\cite{RESIDE} (RESIDE-OUT hereafter), we sample the atmospheric light $A$ and $\beta$ uniformly at random between $[0.7,1.0]$ and $[0.6,1.8]$ respectively. Figure \ref{fig: pipeline}(b) illustrates this process. 

\subsection{Dataset Design}
We now highlight the design choices in constructing our dataset that makes the generated training pairs photorealistic. 
This increase in photorealism is a key that enables training existing dehazing models with superior performance.\medskip 


\noindent \textbf{Scene Selection}.
Unreal's development community offers a wealth of high-quality 3D environment models for photorealistic rendering. These assets are free or can be purchased through the Unreal Marketplace, and can be used to render our training data. Given that haze typically happens in urban scenes, we focus on the generation of clean training images from city scenes. An urban building pack with modular assets and a game-ready high-resolution example scene is used. Since this package is limited in terms of vegetation diversity, a portion of our dataset is rendered using the \textit{City Park Environment Collection}, which contains a demonstration scene of a city park with a wide variety of vegetation.\medskip       

\noindent  \textbf{High-quality Texture and Sky Rendering}.
A critical design consideration of our dataset is photorealism. Our rendering uses textures with a resolution more than 2048$\times$2048 --- 4 times higher than that of the DLSU~\cite{DLSU}. This leads to significantly increased overall photorealism. 

Further, we noticed that images in the DLSU \cite{DLSU} dataset often have monotonous sky patterns (e.g., a white or blue sky with a few clouds) and suffer from noticeable visual artifacts in sky regions (see Figure~\ref{fig:VisualComparison}). To address this issue, we use the skydome in Unreal Engine to render sky regions. Skydome implements a physically-based sky and atmosphere rendering system with HDR sky textures. In doing so, sky regions in our rendered images show major improvement in realism.

\subsection{Training Dehazing Models}
In this section, we described models considered in our experiments, as well as how these models are trained.\medskip


\noindent \textbf{Model Selection}. To benchmark the proposed dataset, we select two latest deep models for single-image haze removal, covering a wide range of design choices of existing dehazing algorithms. The first model, GridDehazeNet~\cite{liu2019griddehazenet}, considers a convolutional neural network (CNN) and directly predicts a haze-free clean image based on a hazy input image. GridDehazeNet represents end-to-end models~\cite{qin2020ffa,liu2019griddehazenet,4KID}. The second model, DehazeFormer \cite{dehazeformer}, adopts a Transformer architecture, predicts quantities related to the global atmospheric light ($A$ in Eq.\ \ref{ASM_equation}) and the transmission map ($t$ in Eq.\ \ref{ASM_equation}), and then uses ASM reformulated with these quantities to compute a clean image. DehazeFormer represents a family of methods incorporating ASM implicitly in the modeling~\cite{AOD_net_ICCV,dehazeformer}. Collectively, GridDehazeNet and DehazeFormer cover both CNN and Transformer architectures.\medskip





\noindent \textbf{Training Dehazing Model with Synthesized Pairs}. Denote the clean image as $I$, and the synthesized hazy image as $H$. Training a deep dehazing network minimizes a certain loss function (often using stochastic gradient descent), given by
\begin{equation}
    \min_{\theta}{\mathbcal{L}[(I,\mathbcal{f}_{\theta}(H)]},
\end{equation}
where $\mathbcal{L}$ is the loss fuction, $\mathbcal{f}$ is the dehazing model and $\theta$ is the parameter of the model. Unlike in \cite{DLSU}, we do not perform domain adaptation, nor do we consider heavy customized built-in physical priors in the deep model. 

\section{Experiments}


 \begin{figure*}[t!]
    \centering
    \includegraphics[width=500pt]{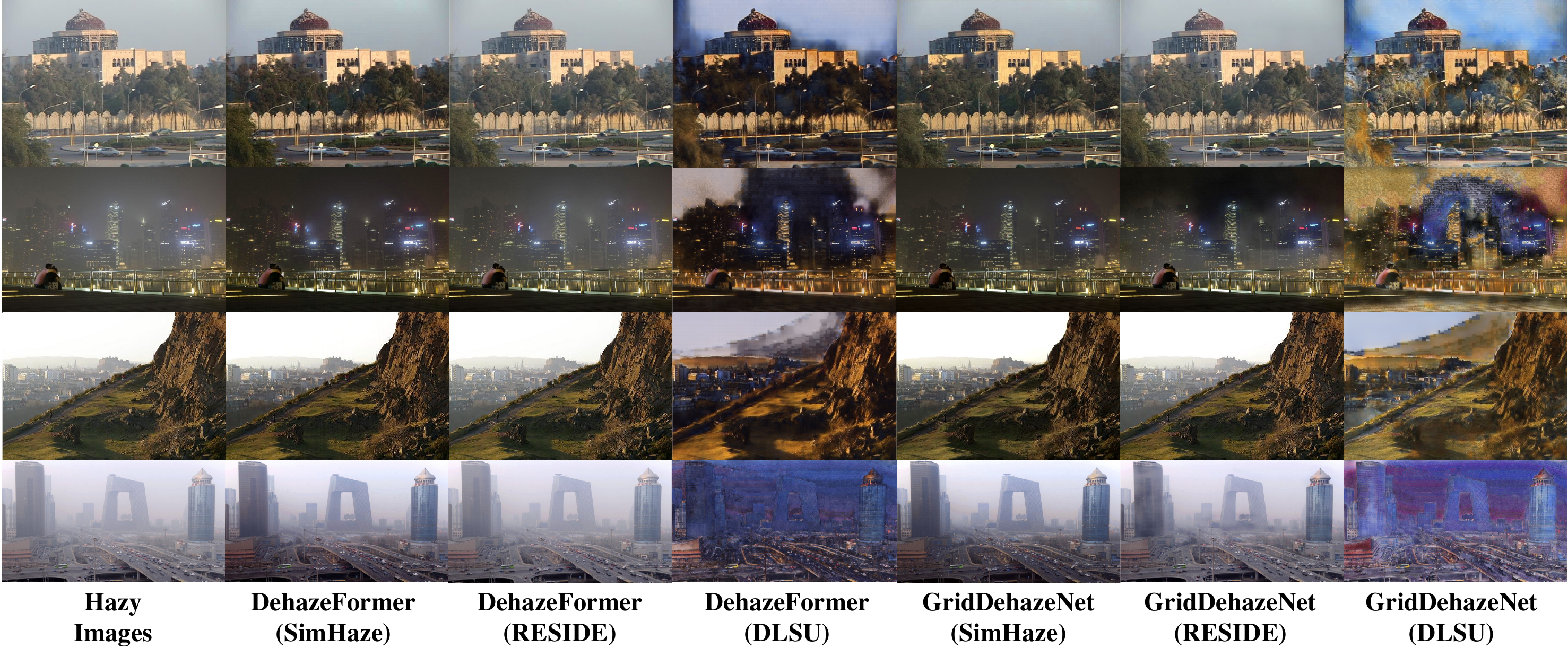}
    \caption{Comparison of dehazing results with different models trained on \textit{SimHaze}, RESIDE\hyp{}OUT, and DLSU and tested on natural images from RTTS and Image2weather. The first column is the hazy images for reference. The 2nd to 4th columns are dehazed images of the DehazeFormer\hyp{}m model trained on three datasets. The reminding columns are results from GridDehazeNet.}
    \label{fig:VisualComparison}
\end{figure*}
For all our experiments, we train those deep models on various datasets (including ours), and compare qualitative and quantitative results on real-world images.\medskip 

\noindent \textbf{Training Datasets}. Models are trained on one of the following datasets: (a) RESIDE-OUT, the outdoor training set of RESIDE (referred to as OTS in some works), where hazy images are synthesized from natural clean images, (b) DLSU \cite{DLSU} dataset, where both clean and hazy images are rendered, and (c) our \textit{SimHaze} with all rendered images. 

RESIDE-OUT is the most popular dataset with real clean images, and DLSU~\cite{DLSU} is our close competitor, which also considers generating data from a game engine.\medskip


\noindent \textbf{Evaluation Protocol}. We conduct qualitative and quantitative evaluations of hazy natural images. Specifically, we use the Real-world Task-driven Testing Set of RESIDE (RTTS) \cite{RESIDE} and the foggy image subset of the Image2weather dataset (Image2weather hereafter) \cite{img2weather}. RTTS is a subset of the RESIDE dataset containing hazy urban scenes. Imag2weather \cite{img2weather} is initially developed for weather condition estimation. The foggy subset contains pictures of foggy cities and supplements RTTS with different backgrounds and fog patterns. 

For quantitative evaluations, We use no-reference image quality assessment scores, as no ground-truth clean images are available for those hazy images. Specifically, we use Fog Density Aware Evaluator (FADE) \cite{FADE} to assess the amount of haze presented in the dehazed images. We also use BRISQUE \cite{BRISQUE}, a well-known no-reference image quality assessment metric, to gauge the image quality of dehazed images. A smaller value of FADE indicates less fog or haze, while a smaller value of BRISQUE suggests a higher image quality.\medskip


\noindent \textbf{Implementation Details}. 
We train GridDehazeNet \cite{liu2019griddehazenet}, and DehazeFormer \cite{dehazeformer} on our \textit{SimHaze}, DLSU and RESIDE-OUT. The same settings are used for all three datasets. For GridDehazeNet, images are cropped to patches of 240$\times$ 240. The mini-batch size is set to 24. The model is trained for ten epochs. Adam \cite{kingma2014adam} is used for optimization.

DehazeFormer has different versions, from DehazeFormer\hyp{}t (tiny) to DehazeFormer\hyp{}m  (middle) for RESIDE\hyp{}OUT. We use DehazeFormer\hyp{}m for the qualitative study and report results on both models for the quantitative study. Images are randomly cropped to 256 $\times$ 256 patches. Mini-batch sizes of 32 and 16 are used for DehazeFormer\hyp{}t and DehazeFormer-m, respectively.  The initial learning
rates are set to $2 \times 10^{−4}$ and $2 \times 10^{−4}$ and gradually decrease to $4 \times 10^{−6}$ and $2 \times 10^{−6}$ based on the cosine annealing \cite{loshchilov2016sgdr} respectively.  
AdamW optimizer \cite{loshchilov2017decoupled} is used.

\subsection{Comparison with Existing Approaches}
We now present our results, starting with a comparison of visual quality and followed by quantitative analysis.  


 \medskip


\noindent \textbf{Qualitative Results.}\label{qualitative} We first validate that models trained on our dataset can produce more visually pleasing dehazed images. The qualitative results of different models are presented side-by-side in Figure \ref{fig:VisualComparison}. 
As one can see, the models trained on RESIDE \cite{RESIDE} do not perform well in removing haze. The models trained on DLSU dataset are good at removing haze, yet show major artifacts in their results, especially in the sky regions. Additionally, one can observe that the overall style of the dehazed images is cartoon-like. This might be attributed to the lack of photorealism in their generated training data. Finally, models trained on our dataset strike a balance between removing haze and maintaining visual details.\medskip

\noindent \textbf{Quantitative Results.}\label{Quantitative}
Our quantitative results are summarized in Table~\ref{tab:QuantitativeScores}. Note that FADE assesses the amount of haze present in an image, BRISQUE evaluates the visual quality of a given image. A high-quality dehazing result should have lower scores on both metrics.

Compared to models trained on our \textit{SimHaze}, models trained on DLSU have lower FADE, yet higher BRISQUE, indicating compromised visual quality. The lower FADE score can be attributed to the higher haze density used when constructing DLSU dataset.   
Models trained on RESIDE-OUT have high FADE and high BRISQUE.
Models trained with our data strike a balance between FADE and BRISQUE. For example, on RESIDE-RTTS, DehazeFormer-t trained on our data has around $30\%$ better BRISQUE compared to both DLSU and RESIDE-OUT while having about $35\%$ better FADE score.
These results, also confirmed by our qualitative results, suggest that models trained on \textit{SimHaze} produce results that strike a balance between removing haze and preserving visual details.

\begin{table}
    \centering
    \resizebox{\columnwidth}{!}{%
    \begin{tabular}{c|c|c|c|c|c}
    \hline
    \multirow{2}{*}{Model} & \multirow{2}{*}{Training Set} & \multicolumn{2}{|c|}{RESIDE-RTTS} & \multicolumn{2}{|c}{Image2weather}\\
    \cline{3-6} 
&  & FADE$\downarrow$ &  BRISQUE$\downarrow$ & FADE$\downarrow$& BRISQUE$\downarrow$ \\
  \hline
      GridDehazeNet & DLSU & \textbf{0.50} & 28.25 & \textbf{0.46} & 24.56  \\
    GridDehazeNet & RESIDE-OUT & 1.52	&29.73& 1.25 & \textbf{22.24}  \\
     GridDehazeNet & SimHaze & 1.31 & \textbf{26.59} & 0.94 & 22.85 \\
     \hline
      DehazeFormer-m & DLSU &\textbf{0.56} &26.43& \textbf{0.50}  & \textbf{21.31}  \\
    
    DehazeFormer-m & RESIDE-OUT & 1.90 &	34.51  &  1.53& 27.57   \\

    DehazeFormer-m & SimHaze & 1.09& \textbf{22.34} & 0.90 & 21.99 \\
        \hline
    DehazeFormer-t & DLSU &\textbf{0.57} &32.09& \textbf{0.54}&26.06  \\
    DehazeFormer-t & RESIDE-OUT & 1.90 & 33.79 & 1.57&23.03  \\
    DehazeFormer-t & SimHaze & 1.23 & \textbf{23.83}  & 0.92 & \textbf{20.74} \\
    \hline
    \end{tabular}
        }
    \caption{Quantitative Results (FADE and BRISQUE scores) on RTTS and Image2weather-foggy datasets. 
    }
    \label{tab:QuantitativeScores} 
\end{table} 

\section{Conclusion}
In this paper, we present a new synthetic data generation pipeline using a modern game engine for training dehazing models. In contrast to previous works, our approach enables training existing deep models with superior performance using synthetic data along and without domain adaptation techniques~\cite{wang2018deep}. We hope our data generation pipeline and finding can spur new ideas and facilitate future research for single-image haze removal. 



{\small
\bibliographystyle{IEEEbib}
\bibliography{egbib.bib}

\begin{thebibliography}{10}

\bibitem{song2022vision}
Yuda Song, Zhuqing He, Hui Qian, and Xin Du,
\newblock ``Vision transformers for single image dehazing,''
\newblock {\em arXiv preprint arXiv:2204.03883}, 2022.

\bibitem{guo2022dehamer}
Chun-Le Guo, Qixin Yan, Saeed Anwar, Runmin Cong, Wenqi Ren, and Chongyi Li,
\newblock ``Image dehazing transformer with transmission-aware 3d position
  embedding,''
\newblock in {\em Proceedings of the IEEE/CVF Conference on Computer Vision and
  Pattern Recognition (CVPR)}, June 2022, pp. 5812--5820.

\bibitem{cai2016dehazenet}
Bolun Cai, Xiangmin Xu, Kui Jia, Chunmei Qing, and Dacheng Tao,
\newblock ``Dehazenet: An end-to-end system for single image haze removal,''
\newblock {\em IEEE Transactions on Image Processing}, vol. 25, no. 11, pp.
  5187--5198, 2016.

\bibitem{Ren-ECCV-2016}
Wenqi Ren, Si~Liu, Hua Zhang, Jinshan Pan, Xiaochun Cao, and Ming-Hsuan Yang,
\newblock ``Single image dehazing via multi-scale convolutional neural
  networks,''
\newblock in {\em European Conference on Computer Vision}, 2016.

\bibitem{li2017all}
Boyi Li, Xiulian Peng, Zhangyang Wang, Jizheng Xu, and Dan Feng,
\newblock ``An all-in-one network for dehazing and beyond,''
\newblock {\em arXiv preprint arXiv:1707.06543}, 2017.

\bibitem{10.1145/3474085.3475432}
Hongyu Li, Jia Li, Dong Zhao, and Long Xu,
\newblock ``Dehazeflow: Multi-scale conditional flow network for single image
  dehazing,''
\newblock in {\em Proceedings of the 29th ACM International Conference on
  Multimedia}, 2021, pp. 2577--2585.

\bibitem{chen2018gated}
Dongdong Chen, Mingming He, Qingnan Fan, Jing Liao, Liheng Zhang, Dongdong Hou,
  Lu~Yuan, and Gang Hua,
\newblock ``Gated context aggregation network for image dehazing and
  deraining,''
\newblock {\em WACV 2019}, 2018.

\bibitem{qin2020ffa}
Xu~Qin, Zhilin Wang, Yuanchao Bai, Xiaodong Xie, and Huizhu Jia,
\newblock ``Ffa-net: Feature fusion attention network for single image
  dehazing,''
\newblock in {\em Proceedings of the AAAI Conference on Artificial
  Intelligence}, 2020, vol.~34, pp. 11908--11915.

\bibitem{narasimhan2002vision}
Srinivasa~G Narasimhan and Shree~K Nayar,
\newblock ``Vision and the atmosphere,''
\newblock {\em International journal of computer vision}, vol. 48, no. 3, pp.
  233, 2002.

\bibitem{RESIDE}
Boyi Li, Wenqi Ren, Dengpan Fu, Dacheng Tao, Dan Feng, Wenjun Zeng, and
  Zhangyang Wang,
\newblock ``Benchmarking single-image dehazing and beyond,''
\newblock {\em IEEE Transactions on Image Processing}, vol. 28, no. 1, pp.
  492--505, 2019.

\bibitem{4KID}
Zhuoran Zheng, Wenqi Ren, Xiaochun Cao, Xiaobin Hu, Tao Wang, Fenglong Song,
  and Xiuyi Jia,
\newblock ``Ultra-high-definition image dehazing via multi-guided bilateral
  learning,''
\newblock in {\em 2021 IEEE/CVF Conference on Computer Vision and Pattern
  Recognition (CVPR)}, 2021, pp. 16180--16189.

\bibitem{gui2021comprehensive}
Jie Gui, Xiaofeng Cong, Yuan Cao, Wenqi Ren, Jun Zhang, Jing Zhang, and Dacheng
  Tao,
\newblock ``A comprehensive survey on image dehazing based on deep learning,''
\newblock {\em arXiv preprint arXiv:2106.03323}, 2021.

\bibitem{DLSU}
Neil~Patrick {Del Gallego}, Joel Ilao, Macario Cordel, and Conrado Ruiz,
\newblock ``A new approach for training a physics-based dehazing network using
  synthetic images,''
\newblock {\em Signal Processing}, vol. 199, pp. 108631, 2022.

\bibitem{unrealengine}
Epic Games,
\newblock ``Unreal engine 4.27,'' .

\bibitem{liu2015learning}
Fayao Liu, Chunhua Shen, Guosheng Lin, and Ian Reid,
\newblock ``Learning depth from single monocular images using deep
  convolutional neural fields,''
\newblock {\em IEEE transactions on pattern analysis and machine intelligence},
  vol. 38, no. 10, pp. 2024--2039, 2015.

\bibitem{liu2019griddehazenet}
Xiaohong Liu, Yongrui Ma, Zhihao Shi, and Jun Chen,
\newblock ``Griddehazenet: Attention-based multi-scale network for image
  dehazing,''
\newblock in {\em Proceedings of the IEEE/CVF international conference on
  computer vision}, 2019, pp. 7314--7323.

\bibitem{dehazeformer}
Yuda Song, Zhuqing He, Hui Qian, and Xin Du,
\newblock ``Vision transformers for single image dehazing,''
\newblock {\em arXiv preprint arXiv:2204.03883}, 2022.

\bibitem{AOD_net_ICCV}
Boyi Li, Xiulian Peng, Zhangyang Wang, Jizheng Xu, and Dan Feng,
\newblock ``Aod-net: All-in-one dehazing network,''
\newblock in {\em Proceedings of the IEEE International Conference on Computer
  Vision (ICCV)}, Oct 2017.

\bibitem{img2weather}
Wei-Ta Chu, Xiang-You Zheng, and Ding-Shiuan Ding,
\newblock ``Image2weather: A large-scale image dataset for weather property
  estimation,''
\newblock in {\em 2016 IEEE Second International Conference on Multimedia Big
  Data (BigMM)}, 2016, pp. 137--144.

\bibitem{FADE}
Lark~Kwon Choi, Jaehee You, and Alan~Conrad Bovik,
\newblock ``Referenceless prediction of perceptual fog density and perceptual
  image defogging,''
\newblock {\em IEEE Transactions on Image Processing}, vol. 24, no. 11, pp.
  3888--3901, 2015.

\bibitem{BRISQUE}
Anish Mittal, Anush~Krishna Moorthy, and Alan~Conrad Bovik,
\newblock ``No-reference image quality assessment in the spatial domain,''
\newblock {\em IEEE Transactions on image processing}, vol. 21, no. 12, pp.
  4695--4708, 2012.

\bibitem{kingma2014adam}
Diederik~P Kingma and Jimmy Ba,
\newblock ``Adam: A method for stochastic optimization,''
\newblock {\em arXiv preprint arXiv:1412.6980}, 2014.

\bibitem{loshchilov2016sgdr}
Ilya Loshchilov and Frank Hutter,
\newblock ``Sgdr: Stochastic gradient descent with warm restarts,''
\newblock {\em arXiv preprint arXiv:1608.03983}, 2016.

\bibitem{loshchilov2017decoupled}
Ilya Loshchilov and Frank Hutter,
\newblock ``Decoupled weight decay regularization,''
\newblock {\em arXiv preprint arXiv:1711.05101}, 2017.

\bibitem{wang2018deep}
Mei Wang and Weihong Deng,
\newblock ``Deep visual domain adaptation: A survey,''
\newblock {\em Neurocomputing}, vol. 312, pp. 135--153, 2018.

\end{thebibliography}
}
\end{document}